\documentclass[10pt,twocolumn,letterpaper]{article}

\usepackage{iccv}
\usepackage{times}
\usepackage{epsfig}
\usepackage{graphicx}
\usepackage{amsmath}
\usepackage{amssymb}
\usepackage{pifont}

\usepackage[numbers]{natbib}

\usepackage{tabularx}
\usepackage{booktabs}
\usepackage{multirow}
\usepackage{makecell}
\usepackage[caption=false]{subfig}

\usepackage{amsmath,amsfonts,bm}

\def\mbf#1{\mathbf{#1}}

\def\1{\bm{1}}

\def\va{{\bm{a}}}

\def\vc{{\bm{c}}}

\def\vh{{\bm{h}}}

\def\vx{{\bm{x}}}

\DeclareMathAlphabet{\mathsfit}{\encodingdefault}{\sfdefault}{m}{sl}
\SetMathAlphabet{\mathsfit}{bold}{\encodingdefault}{\sfdefault}{bx}{n}

\def\gD{{\mathcal{D}}}
\def\gE{{\mathcal{E}}}

\def\gL{{\mathcal{L}}}
\def\gM{{\mathcal{M}}}

\def\gS{{\mathcal{S}}}

\def\sR{{\mathbb{R}}}

\usepackage{xcolor}
\definecolor{MZcolor}{rgb}{0.0,0.0,0.8}
\definecolor{ARcolor}{rgb}{1.0,0.0,0.0}
\definecolor{FTcolor}{rgb}{0.8,0.8,0.0}

\renewcommand{\paragraph}[1]{\vspace{0.2cm}\noindent\textbf{#1}}
\newcommand{\smallparagraph}[1]{\vspace{0.1cm}\noindent\textbf{#1}}

\usepackage[pagebackref=true,breaklinks=true,letterpaper=true,colorlinks,bookmarks=false]{hyperref}

\iccvfinalcopy 

\def\iccvPaperID{2644} 

\ificcvfinal\fi

\begin{document}


\title{MeshTalk: 3D Face Animation from Speech \\ using  Cross-Modality Disentanglement}

\author{Alexander Richard$^1$ \hfill Michael Zollh\"ofer$^1$ \hfill Yandong Wen$^2$ \hfill Fernando de la Torre$^2$ \hfill Yaser Sheikh$^1$\\
$^1$Facebook Reality Labs \quad $^2$Carnegie Mellon University\\
{\tt\small \{richardalex, zollhoefer, yasers\}@fb.com \quad yandongw@andrew.cmu.edu \quad ftorre@cs.cmu.edu}
}

\maketitle
\ificcvfinal\thispagestyle{empty}\fi

\begin{abstract}
This paper presents a generic method for generating full facial 3D animation from speech.
Existing approaches to audio-driven facial animation exhibit uncanny or static upper face animation, fail to produce accurate and plausible co-articulation or rely on person-specific models that limit their scalability.
To improve upon existing models, we propose a generic audio-driven facial animation approach that achieves highly realistic motion synthesis results for the entire face.
%
%
%
%
%
At the core of our approach is a categorical latent space for facial animation that disentangles audio-correlated and audio-uncorrelated information based on a novel cross-modality loss.
Our approach ensures highly accurate lip motion, while also synthesizing plausible animation of the parts of the face that are uncorrelated to the audio signal, such as eye blinks and eye brow motion.
We demonstrate that our approach outperforms several baselines and obtains state-of-the-art quality both qualitatively and quantitatively.
A perceptual user study demonstrates that our approach is deemed more realistic than the current state-of-the-art in over 75\% of cases.
We recommend watching the supplemental video before reading the paper: \small \url{https://github.com/facebookresearch/meshtalk}
\end{abstract}

\vspace{-0.5cm}
\section{Introduction}
\begin{figure}
    \centering
    \includegraphics[scale=0.36]{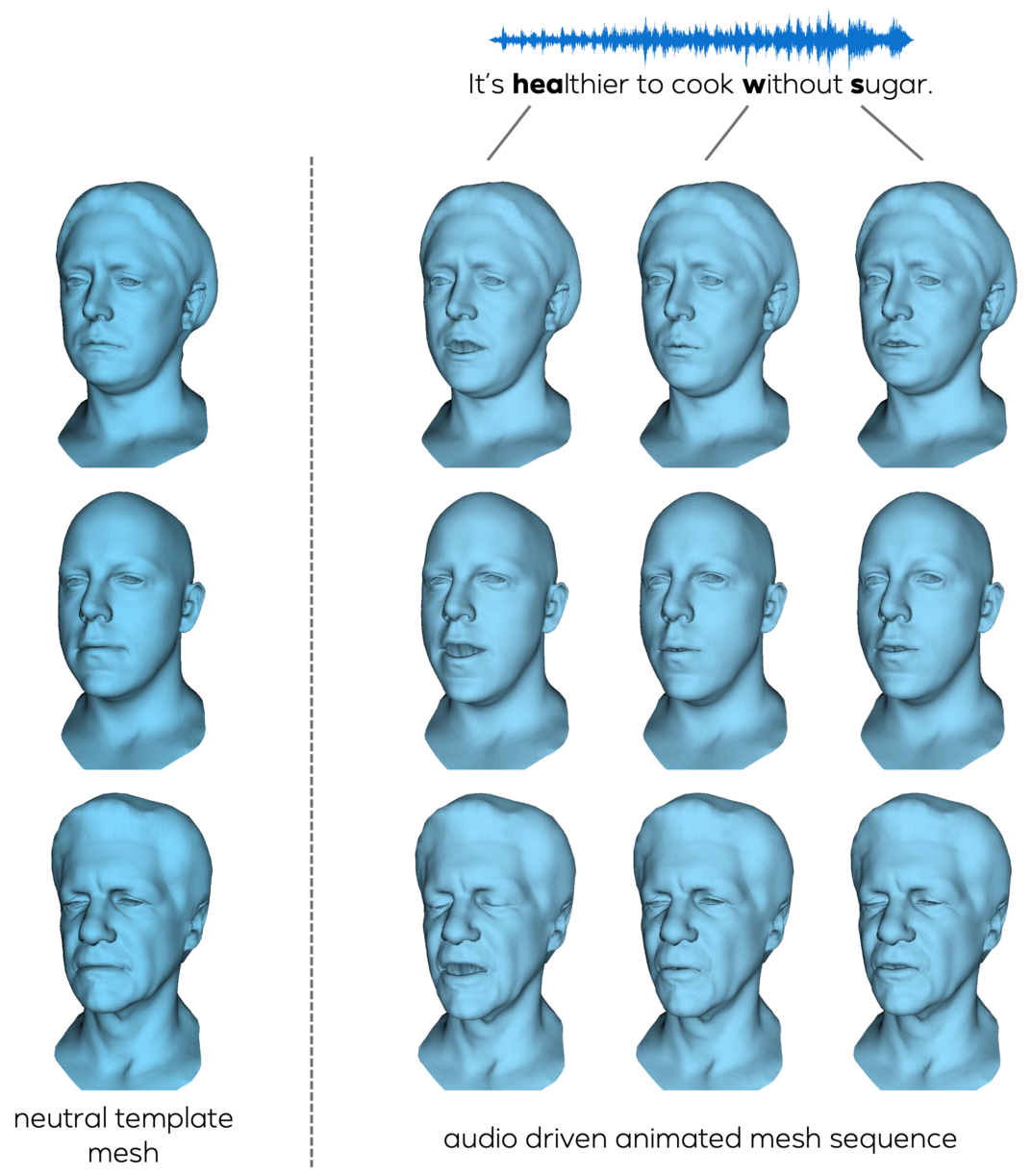}
    \caption{Given a neutral face mesh of a person and a speech signal as input, our approach generates highly realistic face animations with accurate lip shape and realistic upper face motion such as eye blinks and eyebrow raises.}
    \label{fig:teaser}
\end{figure}
Speech-driven facial animation is a highly challenging research problem with several applications such as facial animation for computer games, e-commerce, or immersive VR telepresence.
The demands on speech-driven facial animation differ depending on the application. Applications such as speech therapy or entertainment (e.g., Animojies or AR effects) do not require a very presice level of realism in the animation.
In the production of films, movie  dubbing, driven virtual avatars for e-commerce applications or immersive telepresence, on the contrary, the quality of speech animation requires a high degree of naturalness, plausibility, and has to provide intelligibility comparable to a natural speaker.
The human visual system has been evolutionary adapted to understanding subtle facial motions and expressions.
Thus, a poorly animated face without realistic co-articulation effects or out of lip-sync is deemed to be disturbing for the user.

%

Psychological literature has observed that there is an important degree of dependency between speech and facial gestures.
This dependency has been exploited by audio-driven facial animation methods developed in computer vision and graphics~\cite{bregler1997video,brand1999}.
With the advances in deep learning, recent audio-driven face animation techniques make use of person-specific approaches \cite{nvidia2017siggraph,richard2021audiogaze} that are trained in a supervised fashion based on a large corpus of paired audio and mesh data.
These approaches are able to obtain high-quality lip animation and synthesize plausible upper face motion from audio alone.
To obtain the required training data, high-quality vision-based motion capture of the user is required, which renders these approaches as highly impractical for consumer-facing applications in real world settings.
Recently, Cudeiro \etal~\cite{cudeiro2019capture} extended this work, by proposing a method that is able to generalize across different identities and is thus able to animate arbitrary users based on a given audio stream and a static neutral 3D scan of the user.
While such approaches are more practical in real world settings, they normally exhibit uncanny or static upper face animation~\cite{cudeiro2019capture}.
The reason for this is that audio does not encode all aspects of the facial expressions, thus the audio-driven facial animation problem tries to learn a one-to-many mapping, \ie, there are multiple plausible outputs for every input.
This often leads to over-smoothed results, especially in the regions of the face that are only weakly or even uncorrelated to the audio signal.

%
%
%
%
%
%
%

This paper proposes a novel audio-driven facial animation approach that enables highly realistic motion synthesis for the entire face and also generalizes to unseen identities.
To this end, we learn a novel categorical latent space of facial animation that disentangles audio-correlated and audio-uncorrelated information, \eg, eye closure should not be bound to a specific lip shape.
The latent space is trained based on a novel cross-modality loss that encourages the model to have an accurate upper face reconstruction independent of the audio input and accurate mouth area that only depends on the provided audio input.
This disentangles the motion of the lower and upper face region and prevents over-smoothed results.
Motion synthesis is based on an autoregressive sampling strategy of the audio-conditioned temporal model over the learnt categorical latent space.
Our approach ensures highly accurate lip motion, while also being able to sample plausible animations of parts of the face that are uncorrelated to the audio signal, such as eye blinks and eye brow motion.
%
%
%
%
%
%
%

\section{Related Work}
Speech-based face animation has a long history in computer vision and ranges from artist-friendly stylized and viseme-based models~\cite{jali2016siggraph, visemenet2018siggraph, OculusTechNote} to neural synthesis of 2D~\cite{obama2017siggraph,wiles2018x2face,thies2020nvp} and 3D~\cite{disney2017siggraph,nvidia2017siggraph,richard2021audiogaze} faces.
In the following, we review the most relevant approaches.

\smallparagraph{Viseme-based face animation.}
%
%
In early approaches, a viseme sequence is generated from input text~\cite{ezzat1998miketalk, ezzat2000visual} or directly from speech using HMM-based acoustic models~\cite{verma2003using}.
%
%
Visual synthesis is achieved by blending between templates~\cite{kalberer2001face, kalberer2002speech} or context-dependent viseme models~\cite{martino2006facial}.
%
%
%
%
Given the success of JALI~\cite{jali2016siggraph}, an animator-centric audio-drivable jaw and lip model, Zhou \etal~\cite{visemenet2018siggraph} propose an LSTM-based, near real-time approach to drive a lower face lip model.
Due to their generic nature and artist-friendly design, viseme based approaches are popular for commercial applications particularly in virtual reality~\cite{AmazonSumerian,OculusTechNote}.

\smallparagraph{Speech-driven 2D talking heads.}
Early work on 2D talking heads replaced the problem of learning by searching in existing video for similar utterances as the new speech~\cite{bregler1997video}. 
Brand \etal~\cite{brand1999voice} proposed a generic ML model to drive a facial control model that incorporates vocal and facial dynamic effects such as co-articulation.
The approach of Suwajanakorn \etal \cite{obama2017siggraph} is able to generate video of a single person with accurate lip sync by synthesizing matching mouth textures and compositing them on top of a target video clip.
%
%
Wav2lip \cite{prajwal2020wav2lip} tackles the problem of visual dubbing, while
%
neural Voice Puppetry \cite{thies2020nvp} performs audio-driven facial video synthesis via neural rendering to generate photo-realistic output frames.
X2Face \cite{wiles2018x2face} is an encoder/decoder approach for 2D face animation, \eg, from audio, that can be trained fully self-supervised using a large collection of videos.
Other talking face video techniques \cite{song2019talkingface, chung2017bmvc, vougioukas2018bmvc, zhou2019talking} are based on generative adversarial networks (GANs).
%
%
%
%
%
All the described 2D approaches operate in pixel space and can not be easily generalized to 3D.

\smallparagraph{Speech-driven 3D models.}
Approaches to drive 3D face models mostly use visual input.
While earlier works map from motion captures or 2D video to 3D blendshape models~\cite{deng2006animating, wang2012high, garrido2016reconstruction}, more recent works provide solutions to animate photo-realistic 3D avatars using sensors on a VR headset~\cite{lombardi2018deep, richard2021audiogaze, wei2019siggraph, chu2020expressive}.
These approaches achieve highly realistic results, but they are typically personalized and are not audio-driven.
Most fully speech-driven 3D face animation techniques require either personalized models~\cite{cao2005expressive, nvidia2017siggraph, richard2021audiogaze} or map to lower fidelity blendshape models~\cite{pham2018facefromspeech} or facial landmarks~\cite{eskimez2018iva, greenwook2018interspeech}.
Cao \etal~\cite{cao2005expressive} propose speech-driven animation of a realistic textured personalized 3D face model that requires mocap data from the person to be animated, offline processing and blending of motion snippets.
The fully speech-driven approach of Richard~\etal~\cite{richard2021audiogaze} enables real-time photo-realistic avatars, but is personalized and relies on hours of training data from a single subject.
Karras \etal~\cite{nvidia2017siggraph} learn a speech-driven 3D face mesh from as little as 3-5 minutes of data per subject and condition their model on emotion states that lead to facial expressions.
In contrast to our approach, however, this model has lower fidelity lip sync and upper face expressions, and does not generalize to new subjects.
In~\cite{disney2017siggraph}, a single-speaker model is generalized via re-targeting techniques to arbitrary stylized avatars.
Most closely related to our approach is VOCA~\cite{cudeiro2019capture}, which allows to animate arbitrary neutral face meshes from audio and achieves convincing lip synchronization.
While generating appealing lip motion, their model can not synthesize upper face motion and tends to generate muted expressions.
Moreover, the approach expects a training identity as conditional input to the model.
As shown by the authors in their supplemental video, this identity code has a high impact on the quality of the generated lip synchronization.
Consequentially, we found VOCA to struggle on large scale datasets with hundreds of training subjects.
In contrast to the discussed works, our approach is non-personalized, generates realistic upper face motion, and leads to highly accurate lip synchronization.

\section{Method}

Our goal is to animate an arbitrary neutral face mesh using only speech.
Since speech does not encode all aspects of the facial expressions -- eye-blinks are a simple example of uncorrelated expressive information -- most existing audio-driven approaches exhibit uncanny or static upper face animation~\cite{cudeiro2019capture}.
To overcome this issue, we learn a categorical latent space for facial expressions.
At inference time, an autoregressive sampling from a speech-conditioned temporal model over this latent space ensures accurate lip motion while synthesizing plausible animation of face parts that are uncorrelated to speech.
With this in mind, the latent space should have the following properties:\\
\textit{Categorical.} Most successful temporal models operate on categorical spaces~\cite{vandenoord2016pixelcnn,vandenoord2016wavenet,vaswani2017attention}. In order to use such models, the latent expression space should be categorical as well.\\
\textit{Expressive.} The latent space must be capable of encoding diverse facial expressions, including sparse events like eye blinks.\\
\textit{Semantically disentangled.} Speech-correlated and speech-uncorrelated information should be at least partially disentangled, \eg, eye closure should not be bound to a specific lip shape.

\subsection{Modeling and Learning the Expression Space}

\begin{figure}
    \centering
    \includegraphics[scale=0.19]{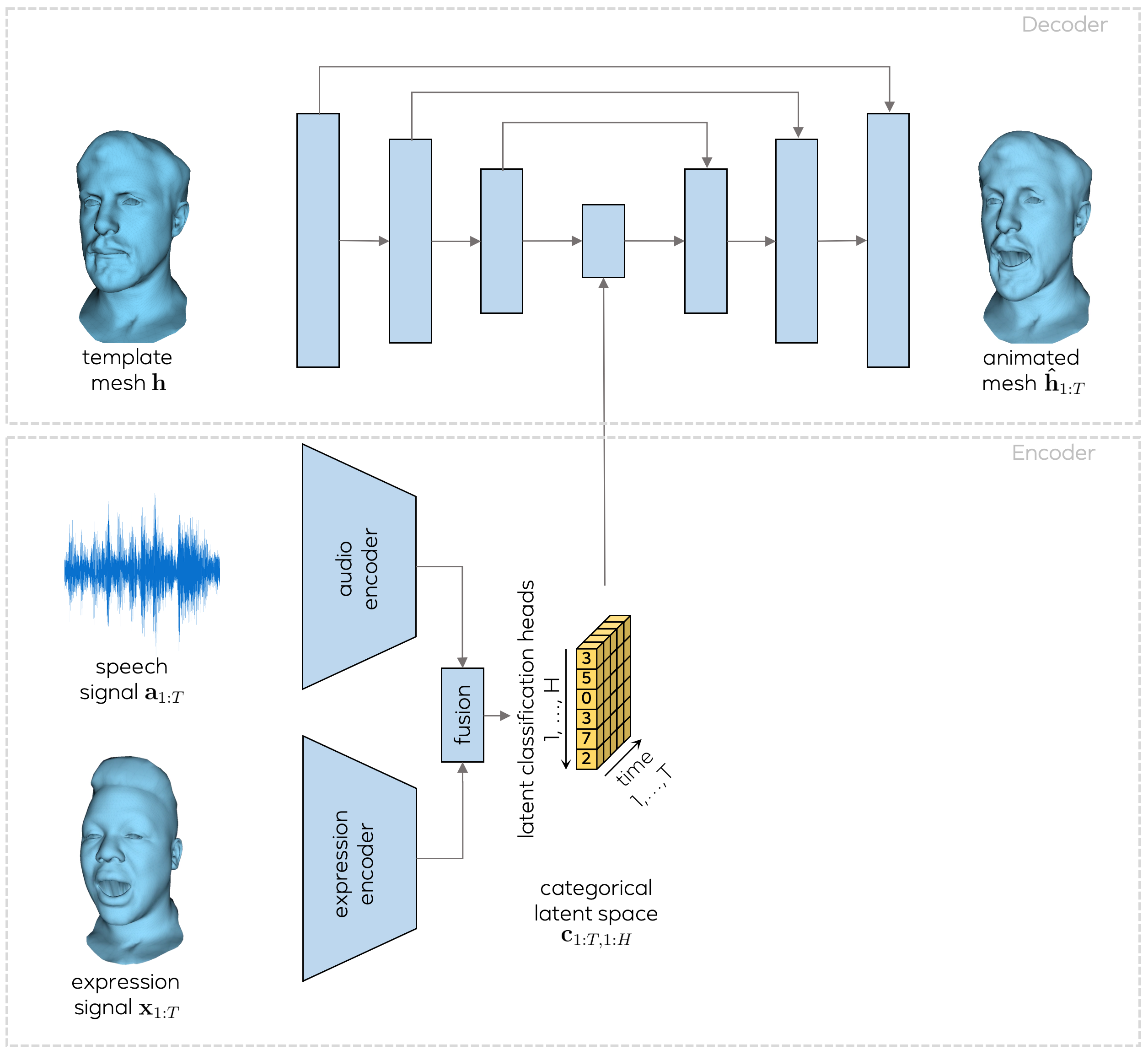}
    \caption{System overview. A sequence of animated face meshes (the expression signal) and a speech signal are mapped to a categorical latent expression space. A UNet-style decoder is then used to animate a given neutral-face template mesh according to the encoded expressions.}
    \label{fig:overview}
    \vspace{-0.3cm}
\end{figure}

Let $ \vx_{1:T} = (\vx_1, \dots, \vx_T), \vx_t \in \sR^{V \times 3} $ be a sequence of $ T $ face meshes, each represented by $ V $ vertices.
Let further $ \va_{1:T} = (\va_1, \dots, \va_T), \va_t \in \sR^D $ be a sequence of $ T $ speech snippets, each with $ D $ samples, aligned to the corresponding (visual) frame $ t $.
Moreover, we denote the template mesh that is required as input by $ \vh \in \sR^{V \times 3} $.

In order to achieve high expressiveness of the categorical latent space, the space must be sufficiently large.
Since this leads to an infeasibly large number of categories $ C $ for a single latent categorical layer, we model $ H $ latent classification heads of $ C $-way categoricals, allowing for a large expression space with a comparably small number of categories as the number of configurations of the latent space is $ C^H $ and therefore grows exponentially in $ H $.
Throughout this paper, we use $ C=128 $ and $ H=64 $.

The mapping from expression and audio input signals to the multi-head categorical latent space is realized by an encoder $ \tilde \gE $ which maps from the space of all audio sequences and all expression sequences (\ie, sequences of animated face meshes) to a $ T \times H \times C $-dimensional encoding
\begin{align}
    \mathrm{\mathbf{enc}}_{1:T, 1:H, 1:C} = \tilde \gE(\vx_{1:T}, \va_{1:T}) \in \sR^{T \times H \times C}.
    \label{eq:encoder}
\end{align}
This continuous-valued encoding is then transformed into a categorical representation using a Gumbel-softmax~\cite{jang2017gumbel} over each latent classification head,
\begin{align}
    \vc_{1:T, 1:H} = \Big[ \mathrm{Gumbel}\big( \mathrm{\mathbf{enc}}_{t, h, 1:C} \big) \Big]_{1:T, 1:H}
\end{align}
such that each categorical component at time step $ t $ and in the latent classification head $ h $ gets assigned one of $ C $ categorical labels, $ c_{t,h} \in \{1,\dots,C\} $.
We denote the complete encoding function, \ie, $ \tilde \gE $ followed by categorization, as $ \gE $.
%

The animation of the input template mesh $ \vh $ is realized by a decoder $ \gD $,
\begin{align}
    \hat \vh_{1:T} = \gD(\vh, \vc_{1:T,1:H}),
    \label{eq:decoder}
\end{align}
which maps the encoded expression onto the provided template $ \vh $.
It thereby generates an animated sequence $ \hat \vh_{1:T} $ of face meshes that looks like the person represented by template $ \vh $, but moves according to the expression code $ \vc_{1:T,1:H} $.
See Figure~\ref{fig:overview} for an overview.

\paragraph{Learning the Latent Space.}
At training time, ground-truth correspondences are only available for the case where \textit{(a)} template mesh, speech signal, and expression signal are from the same identity, and \textit{(b)} the desired decoder output $ \hat \vh_{1:T} $ is equal to the expression input $ \vx_{1:T} $.
Consequentially, training with a simple $ \ell_2 $ reconstruction loss between $ \hat \vh_{1:T} $ and $ \vx_{1:T} $ would lead to the audio input being completely ignored as the expression signal already contains all information necessary for perfect reconstruction -- a problem that leads to poor speech-to-lip synchronization, as we show in Section~\ref{sec:latent_eval}.

We therefore propose a cross-modality loss that ensures information from both input modalities is utilized in the latent space.
Let $ \vx_{1:T} $ and $ \va_{1:T} $ be a given expression and speech sequence.
Let further $ \vh_x $ denote the template mesh for the person represented in the signal $ \vx_{1:T} $.
%
%
Instead of a single reconstruction $ \hat \vh_{1:T} $, we generate two different reconstructions
\begin{align}
    \hat \vh_{1:T}^\text{(audio)} &= \gD\big( \vh_x, \gE(\tilde \vx_{1:T}, \va_{1:T}) \big) \quad \text{and} \\
    \hat \vh_{1:T}^\text{(expr)}  &= \gD\big( \vh_x, \gE(\vx_{1:T}, \tilde \va_{1:T}) \big),
\end{align}
where $ \tilde \vx_{1:T} $ and $ \tilde \va_{1:T} $ are a randomly sampled expression and audio sequence from the training set.
%
In other words, $ \hat \vh_{1:T}^\text{(audio)} $ is a reconstruction given the correct audio but a random expression sequence and $ \hat \vh_{1:T}^\text{(expr)} $ is a reconstruction given the correct expression sequence but random audio.
Our novel cross-modality loss is then defined as
\begin{align}
    \gL_\text{xMod} =& \sum_{t=1}^T \sum_{v=1}^V \gM_v^\text{(upper)}\big( \|\hat h_{t,v}^\text{(expr)} - x_{t,v}\|^2 \big) + \nonumber \\
                     & \sum_{t=1}^T \sum_{v=1}^V \gM_v^\text{(mouth)}\big( \|\hat h_{t,v}^\text{(audio)} - x_{t,v}\|^2 \big),
\end{align}
where $ \gM^\text{(upper)} $ is a mask that assigns a high weight to vertices on the upper face and a low weight to vertices around the mouth.
Similarly, $ \gM^\text{(mouth)} $ assigns a high weight to vertices around the mouth and a low weight to other vertices.

The cross-modality loss encourages the model to have an accurate upper face reconstruction independent of the audio input and, accordingly, to have an accurate reconstruction of the mouth area based on audio independent of the expression sequence that is provided.
Since eye blinks are quick and sparse events that affect only a few vertices, we also found it crucial to emphasize the loss on the eye lid vertices during training.
We therefore add a specific eye lid loss
\begin{align}
    \gL_\text{eyelid} = \sum_{t=1}^T \sum_{v=1}^V \gM_v^\text{(eyelid)}\big( \|\hat h_{t,v} - x_{t,v}\|^2 \big),
\end{align}
where $ \gM^\text{(eyelid)} $ is a binary mask with ones for eye lid vertices and zeros for all other vertices.
The final loss we optimize is $ \gL = \gL_{xMod} + \gL_\text{eyelid} $.
We found that an equal weighting of the two terms works well in practice.

\paragraph{Network Architectures.}
The audio encoder is a four-layer 1D temporal convolutional network similar to the one used in~\cite{richard2021audiogaze}.
The expression encoder has three fully connected layers followed by a single LSTM layer to capture temporal dependencies.
The fusion module is a three-layer MLP.
The decoder $ \gD $ has a UNet-style architecture with additive skip connections, see Figure~\ref{fig:overview}.
This architectural inductive bias prevents the network from diverging from the given template mesh too much.
In the bottleneck layer, the expression code $ \vc_{1:T,1:H} $ is concatenated with the encoded template mesh.
The bottleneck layer is followed by two LSTM layers to model temporal dependencies between frames followed by three fully connected layers remapping the representation to vertex space.

See the supplementary material for more details.

\subsection{Audio-Conditioned Autoregressive Modeling}
\begin{figure}
    \centering
    \includegraphics[scale=0.25]{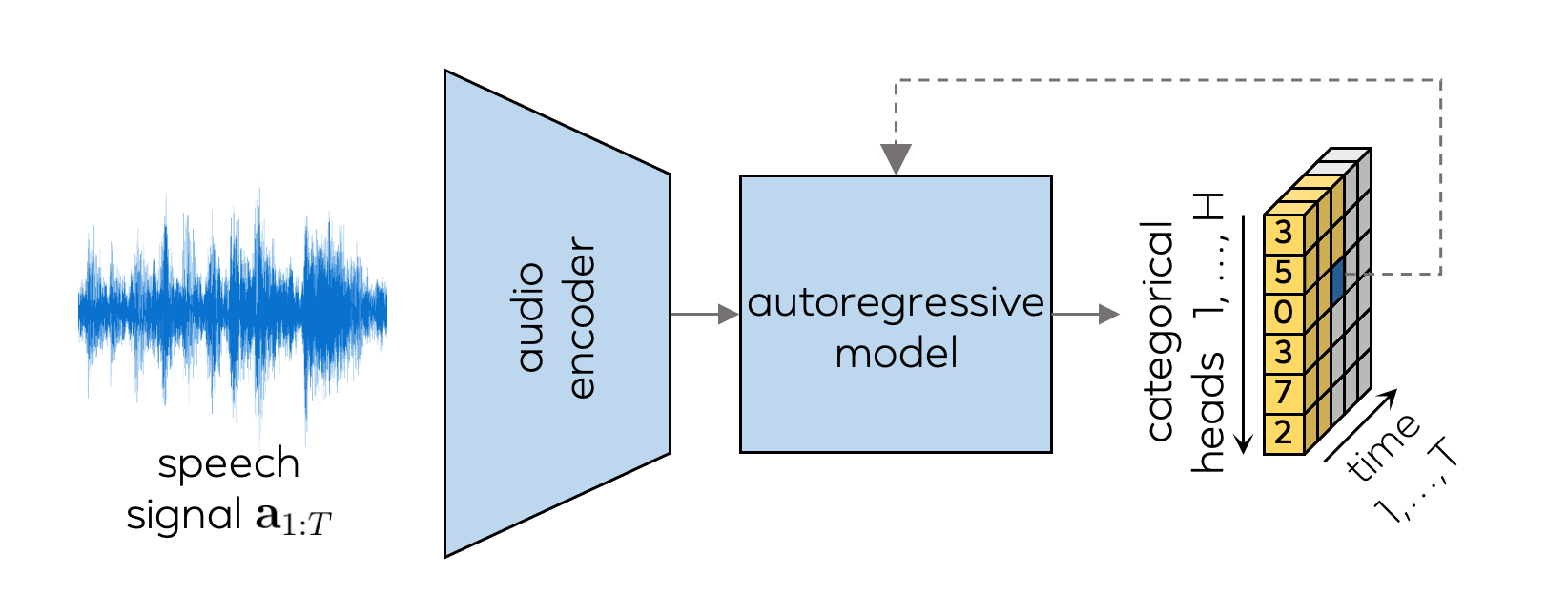}
    \caption{Autoregressive model. Audio-conditioned latent codes are sampled for each position $ c_{t,h} $ in the latent expression space, where the model only has access to previously generated labels as defined in Equation~\eqref{eq:autoregressive}.}
    \label{fig:autoregressive}
\end{figure}
When driving a template mesh using audio input alone, the expression input $ \vx_{1:T} $ is not available.
With only one modality given, missing information that can not be inferred from audio has to be synthesized.
Therefore, we learn an autoregressive temporal model over the categorical latent space.
This model allows to sample a latent sequence that generates plausible expressions and is consistent with the audio input.
Following Bayes' Rule, the probability of a latent embedding $ \vc_{1:T,1:H} $ given the audio input $ \va_{1:T} $ can be decomposed as
\begin{align}
    p(\vc_{1:T,1:H}|\va_{1:T}) = \prod_{t=1}^T \prod_{h=1}^H p(c_{t,h}|\vc_{<t,1:H}, \vc_{t,<h}, \va_{\leq t}).
    \label{eq:autoregressive}
\end{align}
Note that we assumed temporal causality in the decomposition, \ie, a category $ c_{t,h} $ at time $ t $ only depends on current and past audio information $ \va_{\leq t} $ rather than on past and future context $ \va_{1:T} $.
We model this quantity with an autoregressive convolutional network similar to PixelCNN~\cite{vandenoord2016pixelcnn}.
Our autoregressive temporal CNN has four convolutional layers with increasing dilation along the temporal axis.
The convolutions are masked such that for the prediction of $ c_{t,h} $ the model only has access to information from all categorical heads in the past, $ \vc_{<t,1:H} $, and the preceding categorical heads at the current time step, $ \vc_{t,<h} $, see yellow blocks in Figure~\ref{fig:autoregressive}.
To train the autoregressive model, we generate training data using the pretrained encoder $ \gE $, which maps the expression and audio sequences $ (\vx_{1:T}, \va_{1:T}) $ in the training set to their categorical embeddings, see Equation~\eqref{eq:encoder}.
The autoregressive model is then optimized on these correspondences using teacher forcing and a cross-entropy loss over the latent categorical labels.
At inference time, a categorical expression code is sequentially sampled for each position $ c_{t,h} $ using the trained audio-conditioned autoregressive network.

\section{Evaluation}
\smallparagraph{Dataset.}
Existing works are typically trained on less than a dozen subjects~\cite{nvidia2017siggraph, disney2017siggraph, richard2021audiogaze} and available datasets are tracked with low fidelity~\cite{cudeiro2019capture}, \eg, not including eye lids, facial hair, or eyebrows, and therefore render unfit to demonstrate high fidelity full-face motion from speech generalizing over arbitrary identities.

In this work, we use an in-house dataset of $250$ subjects, each of which is reading a total of $50$ phonetically balanced sentences.
The sequences are captured at $30$ frames per second and face meshes are tracked from $80$ synchronized cameras surrounding the subject's head.
We use a face model with $6,172$ vertices that has a high level of detail including eye lids, upper face structure, and different hair styles.
In total, our data amounts to $13$h of paired audio-visual data, or $1.4$ million frames of tracked 3D face meshes.
We train our model on the first $40$ sentences of $200$ subjects and use the remaining $10$ sentences of the remaining $50$ subjects as validation ($10$ subjects) and test set ($40$ subjects).
All shown qualitative and quantitative results are obtained using the held-out subjects and sentences in the test set.
We release a subset of $ 16 $ subjects of this dataset and our model using only these subjects as a baseline to compare against.
We refer the reader to the github repository for all necessary information.\footnote{\url{https://github.com/facebookresearch/meshtalk}}

\smallparagraph{Speech Features.}
Our audio data is recorded at $16$kHz.
For each tracked mesh, we compute the Mel spectrogram of a $600$ms audio snippet starting $500$ms before and ending $100$ms after the respective visual frame.
We extract $80$-dimensional Mel spectral features every $10$ms, using $1,024$ frequency bins and a window size of $800$ for the underlying Fourier transform.

\subsection{Disentangling Audio and Expression}
\label{sec:latent_eval}
In this section, we show that our model successfully learns a latent representation that allows to control upper face motion and lip synchronization from different input modalities.
Specifically, we aim to answer these questions:

\noindent\textbf{(1)}
In Figure~\ref{fig:overview}, the encoder maps a sequence of audio features and face meshes to the latent space.
At training time, the expression input $ \vx_{1:T} $ is exactly the target signal that would minimize the loss at the output of the decoder.
\textit{Why is it crucial to have audio as input to learn the latent space?}

\noindent\textbf{(2)}
Many multi-modal approaches suffer from the problem that the weaker modality is ignored~\cite{moe2019Neurips,wang2020whatmakes}.
\textit{How can this effect be avoided, considering that the expression input is already signal-complete?}

\noindent\textbf{(3)}
Given the issues discussed above, it is interesting to investigate the structure of the latent space.
\textit{Are latent space and animated mesh regions semantically disentangled across modalities?}

\noindent\textbf{(4)}
We use a categorical latent space.
\textit{Is this preferable over a continuous latent space?}

\paragraph{(1) Multi-modal encoder inputs.}
\begin{table}
    \footnotesize
    \centering
    \begin{tabularx}{0.48\textwidth}{Xrrr}
        \toprule
                           &                          & reconstruction       & autoregr. model \\
        encoder inputs     & decoder loss             & error (in mm)        & perplexity \\
        \midrule
        expression         & $ \ell_2 $               & $ 1.156 $            & $ 1.853 $ \\
        expr. + audio      & $ \ell_2 $               & $ 1.124 $            & $ 1.879 $ \\
        expr. + audio      & $ \gL_\text{xMod} $      & $ 1.244 $            & $ \mbf{1.669} $ \\
        \bottomrule
    \end{tabularx}
    \caption{Different strategies to learn the latent space and their impact on the autoregressive model.}
    \label{tab:perplexity}
\end{table}
We start with a discussion on learning a latent space from audio and expression inputs.
One straightforward way to learn a latent space is to \textit{omit the audio input} to the encoder in Figure~\ref{fig:overview}.
Limited capacity of the latent space and the inductive bias of the decoder, specifically its skip connections, make sure that even in this case, sufficient information is used from the template geometry (see supplemental video for an example).
Not surprisingly, this setup also leads to a low reconstruction error\footnote{We refer to the reconstruction error of a model as its $ \ell_2 $ error on the test set when all inputs, \ie, template mesh and encoder input, are from the same person.
The autoregressive model is not used for reconstruction.} as shown in Table~\ref{tab:perplexity}.
The major caveat of this strategy, however, is the structure of the latent space.
Since there is no incentive to disentangle information about independent parts of the face, latent representations can have a strong entanglement between eye motion and mouth shape.
As a consequence, the autoregressive model (Equation~\eqref{eq:autoregressive}) fails to generate latent representations of accurate lip shape if it is required to produce temporally consistent and plausible upper face motion at the same time.
Consequentially, lip motion is less accurate and facial motion more muted when compared to our model that uses audio and expression to learn the latent space.

We quantify this effect by evaluating the perplexity of the autoregressive model.
Given a categorical latent representation $ \vc_{1:T,1:H} $ of test set data, the perplexity is
\begin{align}
    PP = p(\vc_{1:T,1:H}|\va_{1:T})^{-\frac{1}{T \cdot H}},
\end{align}
\ie, the inverse geometric average of the likelihood of the latent representations under the autoregressive model.
Intuitively, a low perplexity means that at each prediction step the autoregressive model only has a small number of potential categories to choose from, whereas high perplexity means the model is less certain which categorical representation to choose next.
A perplexity of $1$ would mean the autoregressive model is fully deterministic, \ie, the latent embedding is fully defined by the conditioning audio input.
As there are face motions uncorrelated with audio, this does not happen in practice.
A comparison of row one and three in Table~\ref{tab:perplexity} shows that learning the latent space from audio and expression input leads to a stronger and more confident audio-conditioned autoregressive model than learning the latent space from expression inputs alone.
This observation is consistent with the qualitative evaluation in the supplemental video.

\paragraph{(2) Enforcing the usage of audio input.}
The training loss of the decoder has major impact on how the model makes use of the different input modalities.
Since the expression input is sufficient for exact reconstruction, a simple $ \ell_2 $ loss on the desired output meshes will cause the model to ignore the audio input completely and the results are similar to the above case where no audio was given as encoder input, see Table~\ref{tab:perplexity} row one and two.
The cross-modality loss $ \gL_\text{xMod} $ offers an effective solution to this issue.
The model is encouraged to learn accurate lip shape even if the expression input is exchanged by different, random expressions.
Similarly, upper face motion is encouraged to remain accurate independent of the audio input.
The last row in Table~\ref{tab:perplexity} shows that the cross-modality loss does not negatively affect \textit{expressiveness} of the learnt latent space (\ie, the reconstruction error is small for all latent space variants) but positively affects the autoregressive model's perplexity.

\paragraph{(3) Cross-Modality Disentanglement.}
\begin{figure}
    \centering
    \includegraphics[scale=0.3]{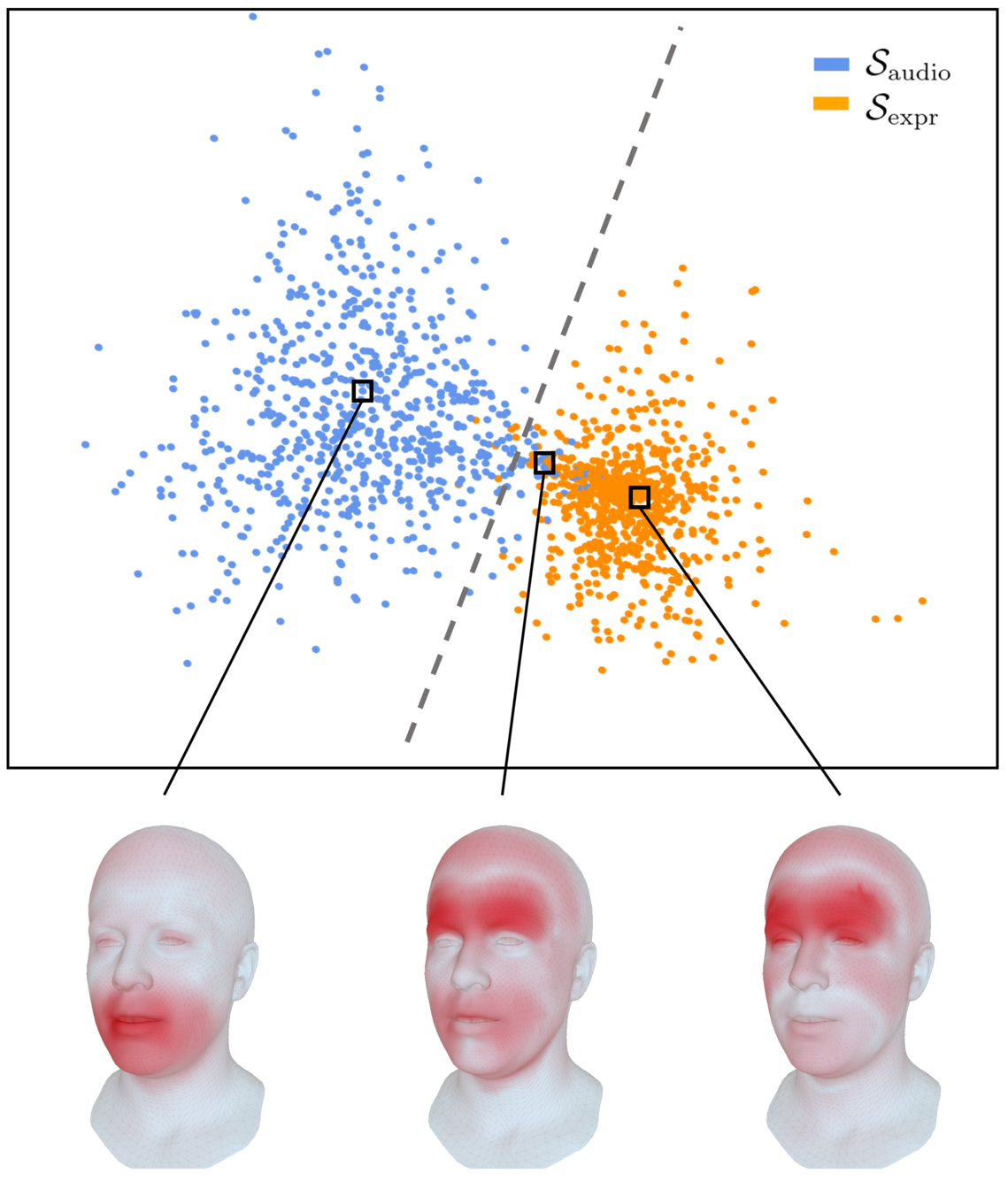}
    \caption{Visualization of the latent space. Latent configurations caused by changes in the audio input are clustered together. Latent configurations caused by changes in the expression input form another cluster. Both clusters can be well separated with minimal leakage into each other.}
    \label{fig:latent_space}
\end{figure}
We demonstrate that the cross-modal disentanglement with its properties analyzed above in fact leads to a structured latent space and that each input modality has different effects on the decoded face meshes.
To this end, we generate two different sets of latent representations, $ \gS_\text{audio} $ and $ \gS_\text{expr} $.
$ \gS_\text{audio} $ contains latent codes obtained by fixing the expression input to the encoder and varying the audio input.
Similarly, $ \gS_\text{expr} $ contains latent codes obtained by fixing the audio input and varying the expression input.
In the extreme case of perfect cross-modal disentanglement, $ \gS_\text{audio} $ and $ \gS_\text{expr} $ form two non-overlapping clusters.
We fit a separating hyper-plane on the points in $ \gS_\text{audio} \cup \gS_\text{expr} $ and visualize a 2D projection of the result in Figure~\ref{fig:latent_space}.
Note that there is only minimal leakage between the clusters formed by $ \gS_\text{audio} $ and $ \gS_\text{expr} $.
Below the plot, we visualize which face vertices are most moved by latent representations within the cluster of $ \gS_\text{audio} $ (left), within the cluster of $ \gS_\text{expr} $ (right), and close to the decision boundary (middle).
While audio mostly controls the mouth area and expression controls the upper face, latent representations close to the decision boundaries influence face vertices in all areas, which suggests that certain upper face expressions are correlated to speech, e.g., raising the eyebrows.

\begin{figure}
    \centering
    \hfill
    \subfloat[Vertices most influenced by audio input.]{
        \hspace{0.05\textwidth}
        \includegraphics[scale=0.075]{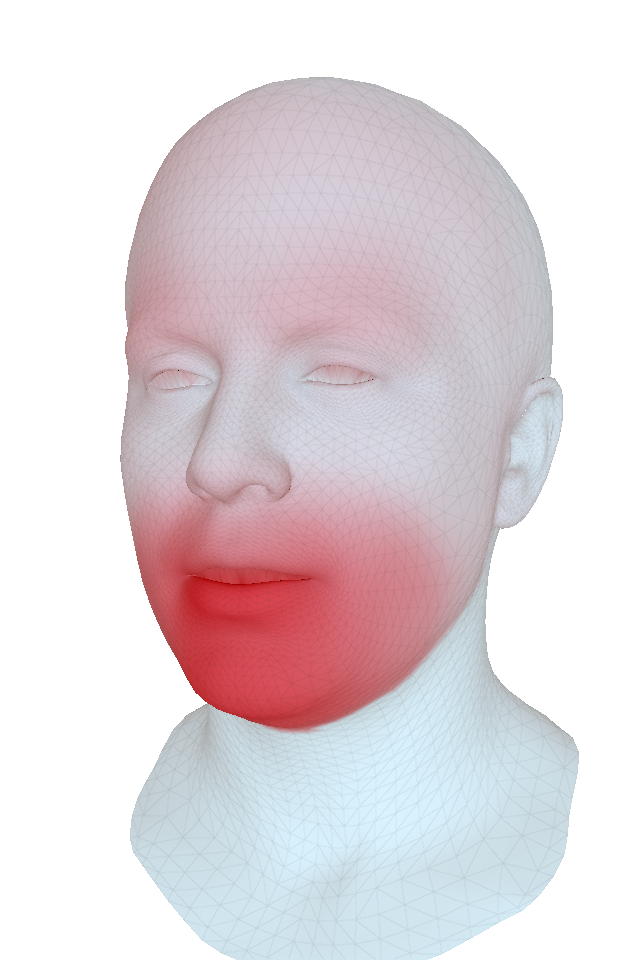}
        \hspace{0.05\textwidth}
        \label{fig:heatmesh_a}
    }
    \hfill
    \subfloat[Vertices most influenced by expression input.]{
        \hspace{0.05\textwidth}
        \includegraphics[scale=0.075]{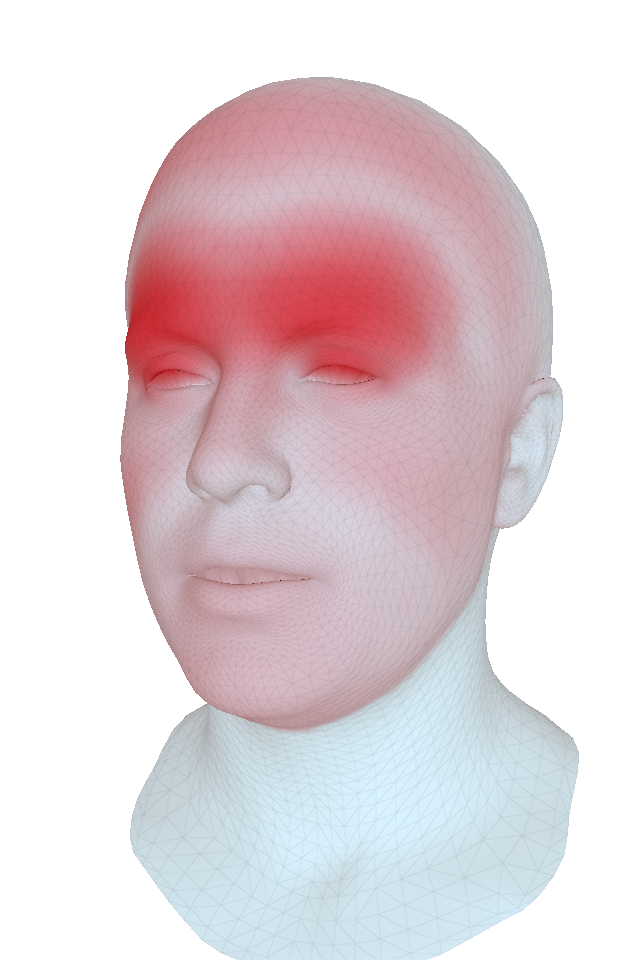}
        \hspace{0.05\textwidth}
        \label{fig:heatmesh_b}
    }
    \hfill
    \vspace{0.1cm}
    \caption{Impact of the audio and expression modalities on the generated face meshes. Audio steers primarily the mouth area but has also a visible impact on eyebrow motion. Expression meshes influence primarily the upper face parts including the eye lids.}
    \label{fig:heatmesh}
\end{figure}
In Figure~\ref{fig:heatmesh}, we show the vertices that are most affected by decoding of all latent representations in $ \gS_\text{audio} $ and $ \gS_\text{expr} $, respectively.
It becomes evident that the loss $ \gL_\text{xMod} $ leads to a clear cross-modality disentanglement into upper and lower face motion.
Yet, it is notable that audio, besides its impact on lips and jaw, has a considerable impact on the eyebrow area (Figure~\ref{fig:heatmesh_a}).
We show in the supplemental video that our model, in fact, learns correlations between speech and eyebrow motion, \eg, raising the eyebrows when words are emphasized.

\paragraph{(4) Necessity of categorical space.}
\begin{table}[tb]
    \centering
    \footnotesize
    \begin{tabularx}{0.45\textwidth}{Xrr}
        \toprule
        latent space    & vertex error (mm)  & lip error (mm)   \\
        \midrule
        continuous      & $ 1.975 $          & $ 4.578 $        \\ 
        categorical     & $ \mathbf{1.244} $ & $ \mathbf{3.184} $        \\
        \bottomrule
    \end{tabularx}
    \caption{Vertex errors for categorical vs.~continuous latent spaces.}
    \label{tab:errors}
\end{table}

\begin{figure}[tb]
    \centering
    \hfill
    \subfloat[Continuous latent space.]{
        \hspace{0.05\textwidth}
        \includegraphics[scale=0.075]{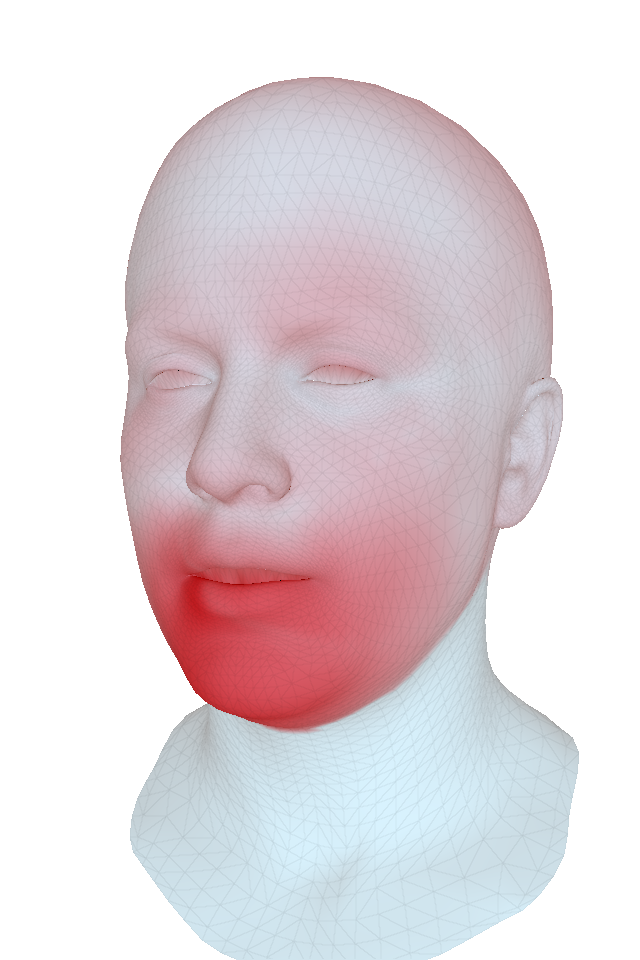}
        \hspace{0.05\textwidth}
        \label{fig:cont_heatmesh}
    }
    \hfill
    \subfloat[Categorical latent space.]{
        \hspace{0.05\textwidth}
        \includegraphics[scale=0.075]{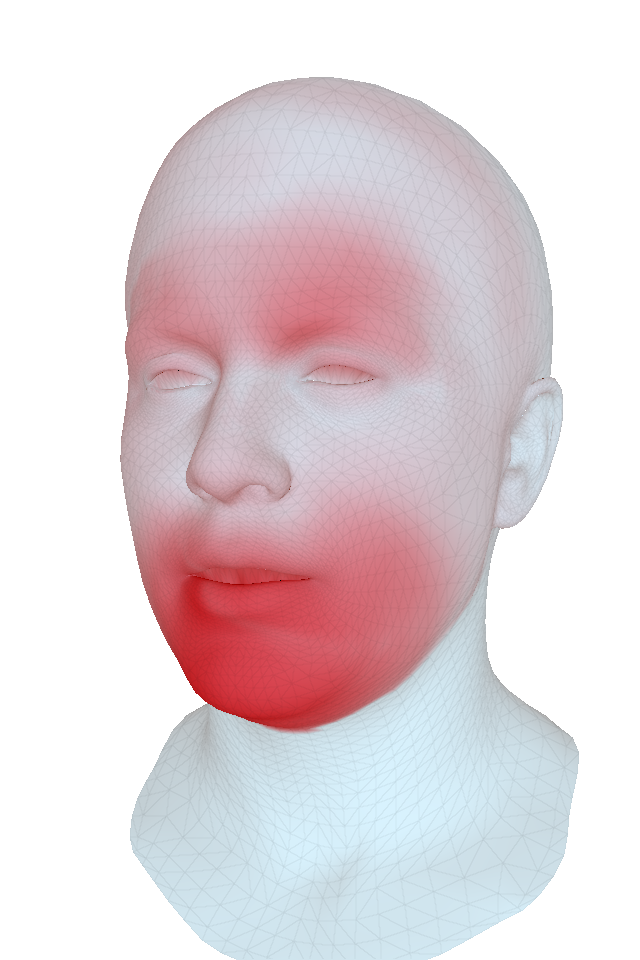}
        \hspace{0.05\textwidth}
        \label{fig:cat_heatmesh}
    }
    \hfill
    \caption{Standard deviation of vertex positions of audio-driven face meshes. The darker the red, the higher the motion.}
    \label{fig:vertex_motion}
\end{figure}
We compared the categorical latent space with a continuous latent space.
To maintain the stochastic property for the continuous case, the encoder predicts a mean and variance for each frame from which a representation is then sampled.
At inference time, the auto-regressive model predicts mean and variance from audio input and all past latent representations. The next latent embedding is then sampled from these mean and variance predictions.
We observe that lip error and overall vertex error is worse than for the categorical latent space (Tab.~\ref{tab:errors}).
In Fig.~\ref{fig:vertex_motion}, we plot the variance of vertices of audio-driven face meshes.
Note how upper face motion collapses towards mean expressions for the continuous space (only little vertex motion, Fig.~\ref{fig:vertex_motion}a), whereas the categorical space allows to sample rich and diverse upper face motion (Fig.~\ref{fig:vertex_motion}b).

\subsection{Audio-Driven Evaluation}

\begin{table}
    \footnotesize
    \centering
    \begin{tabularx}{0.48\textwidth}{Xr}
        \toprule
                                                                    & lip vertex error (in mm) \\
        \midrule
        VOCA~\cite{cudeiro2019capture}                              & $ 3.720 $ \\
        VOCA~\cite{cudeiro2019capture} + our audio encoder          & $ 3.472 $ \\
        Ours                                                        & $ \mbf{3.184} $ \\
        \bottomrule
    \end{tabularx}
    \caption{Lip errors of our approach compared to VOCA.}
    \label{tab:lip_errors}
\end{table}

\smallparagraph{Lip-sync Evaluation.}
We compare our approach to VOCA~\cite{cudeiro2019capture}, the state of the art in audio-driven animation of arbitrary template face meshes.
To evaluate the quality of the generated lip synchronization, we define the lip error of a single frame to be the maximal $ \ell_2 $ error of all lip vertices and report the average over all frames in the test set.
Since upper lip and mouth corners move much less than the lower lip, we found that the average over all lip vertex errors tends to mask inaccurate lip shapes, while the maximal lip vertex error per frame correlates better with the perceptual quality.

For the comparison with VOCA, which expects a conditioning on a training identity, we sample said identity at random.
In addition to the original implementation of VOCA, we also compare to a variant where we replace the DeepSpeech features~\cite{hannun2014deep} used in~\cite{cudeiro2019capture} with Mel spectrograms and our audio encoder.
Table~\ref{tab:lip_errors} shows that our approach achieves a lower lip error per frame on average.
As shown in the supplemental material of~\cite{cudeiro2019capture}, the quality of VOCA is strongly dependent on the chosen conditioning identity.
On a challenging dataset with highly detailed meshes, we find this effect to be amplified even more and observed VOCA to produce much less pronounced lip motion, particularly missing most lip closures.
We provide a side-by-side comparison in the supplemental video.

\smallparagraph{Perceptual Evaluation.}
\begin{table}
    \def\arrvline{\hfil\kern\arraycolsep\vline\kern-\arraycolsep\hfilneg}
    \footnotesize
    \centering
    \begin{tabularx}{0.48\textwidth}{Xrrrr}
        \toprule
                                                       & \multicolumn{3}{c}{favorability}                           & \multirow{2}{*}{\makecell{ours better \\ or equal}} \\
                                                         \cmidrule(lr){2-4}
                                                       & competitor          & equal             & ours             &  \\
        \midrule
        \multicolumn{5}{l}{\textbf{ours vs. VOCA~\cite{cudeiro2019capture}}} \\
        full-face                                      & $ 24.7\% $          & $ 20.9\% $        & $ 54.4\% $      \arrvline & $ \mbf{75.3\%} $ \\
        lip sync                                       & $ 23.0\% $          & $ 19.8\% $        & $ 57.2\% $      \arrvline & $ \mbf{77.0\%} $ \\
        upper face                                     & $ 33.6\% $          & $ 21.6\% $        & $ 44.8\% $      \arrvline & $ \mbf{66.4\%} $ \\
        \midrule
        \multicolumn{5}{l}{\textbf{ours vs. ground truth}} \\
        full-face                                      & $ 42.1\% $          & $ 35.7\% $        & $ 22.2\% $      \arrvline & $ \mbf{57.9\%} $ \\
        lip sync                                       & $ 45.1\% $          & $ 34.1\% $        & $ 20.8\% $      \arrvline & $ \mbf{54.9\%} $ \\
        upper face                                     & $ 68.5\% $          & $  6.9\% $        & $ 24.6\% $      \arrvline & $ \mbf{31.5\%} $ \\
        \bottomrule
    \end{tabularx}
    \caption{Perceptual study. Human participants where asked which of two presented clips are more realistic as full-face clips, upper face only, or in terms of lip sync. For each row, $ 400 $ pairs of side-by-side clips have been ranked by favorability.}
    \label{tab:user_study}
\end{table}
We presented side-by-side clips of our approach versus either VOCA or tracked ground truth to a total of $100$ participants and let them judge three subtasks:
a full face comparison, a lip sync comparison, where we showed only the region between the chin and the nose, and an upper face comparison, where we showed only the face from the nose upwards, see Table~\ref{tab:user_study}.
For each row, $400$ pairs of short clips each containing one sentence spoken by a subject from the test set have been evaluated.
Participants could choose to either favor one clip over the other or rank them both as equally good.

In a comparison against VOCA, our approach has been ranked better or equal in more than $75\%$ of the cases for full face animation and lip sync and is ranked better or equal than VOCA in $77\%$ of the cases.
For upper face motion alone, the results are slightly lower but participants still heavily favored our approach ($66.4\%$ of cases better than or equal to VOCA).
Note that most frequently our approach has been strictly preferred over VOCA; equal favorability only makes up a minority of cases.
When comparing our approach to tracked ground truth, most participants found ground truth favorable.
Yet, with $ 35.7\% $ of full-face examples being ranked equally good and only $ 42.1\% $ ranked less favorable than ground truth, our approach proves to be a strong competitor.

\smallparagraph{Qualitative Examples.}
Figure~\ref{fig:teaser} shows audio-driven examples generated with our approach.
The lip shapes are consistent with the respective speech parts among all subjects, while unique and diverse upper face motion such as eye brow raises and eye blinks are generated separately for each sequence.
Further examples generated with our approach and comparisons to other models can be found in the supplemental video.

\subsection{Re-Targeting}
\begin{figure}
    \centering
    \includegraphics[scale=0.36]{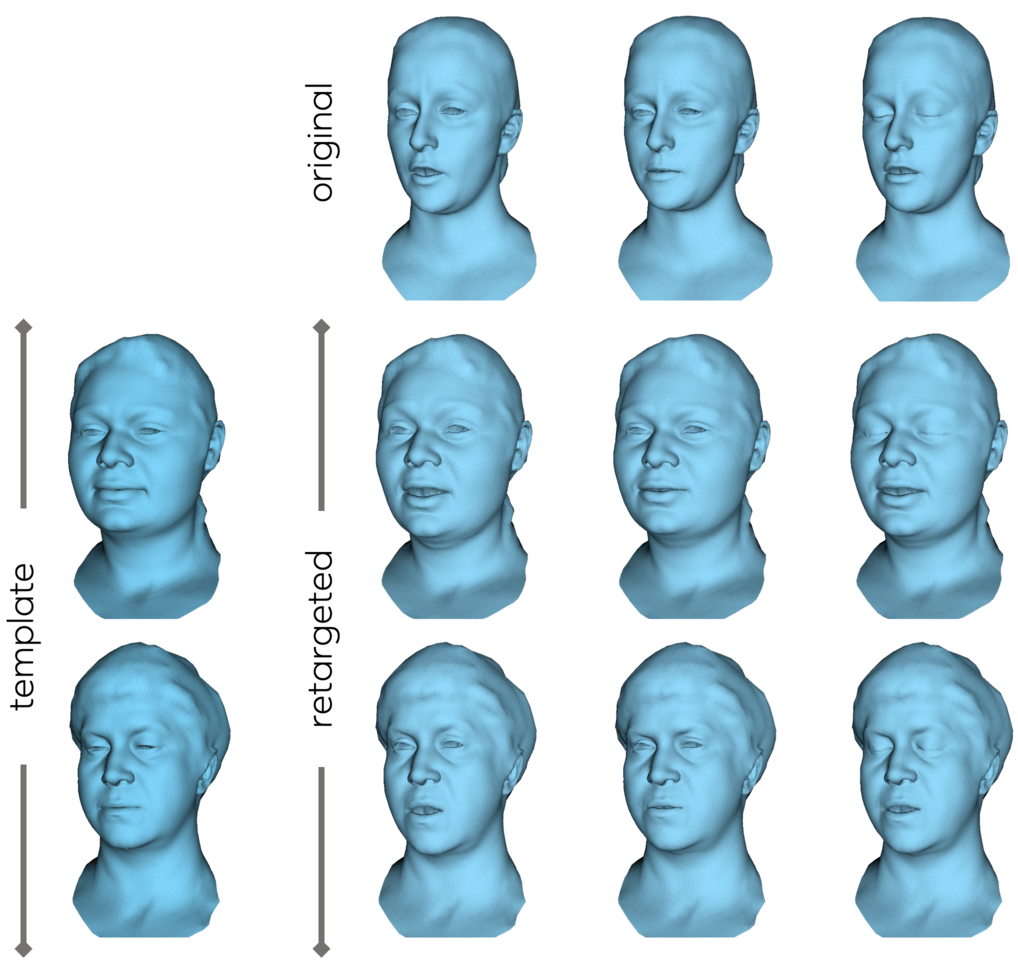}
    \caption{Re-targeting. Given an animated mesh and neutral templates of other identities, our approach accurately re-targets facial expressions such as lip shape, eye closure, and eyebrow raises.}
    \label{fig:retargeting}
\end{figure}
Re-targeting is the process of mapping facial motion from one identity's face onto another identity's face.
Typical applications are movies or computer games, where an actor animates a face that is not his own.
Our system can successfully address this problem, as we demonstrate in the supplemental video.
Using the architecture displayed in Figure~\ref{fig:overview}, the latent expression embedding is computed from the original (tracked) face mesh and the audio of the source identity.
The template mesh is of the desired target identity.
Our system maps the audio and original animated face mesh to a latent code and decodes it to an animated version of the template mesh.
Note that no autoregressive modeling is required for this task.
Besides the video, Figure~\ref{fig:retargeting} shows examples of re-targeted facial expressions from one identity to another.

\subsection{Mesh Dubbing}
\begin{figure}
    \centering
    \includegraphics[scale=0.22]{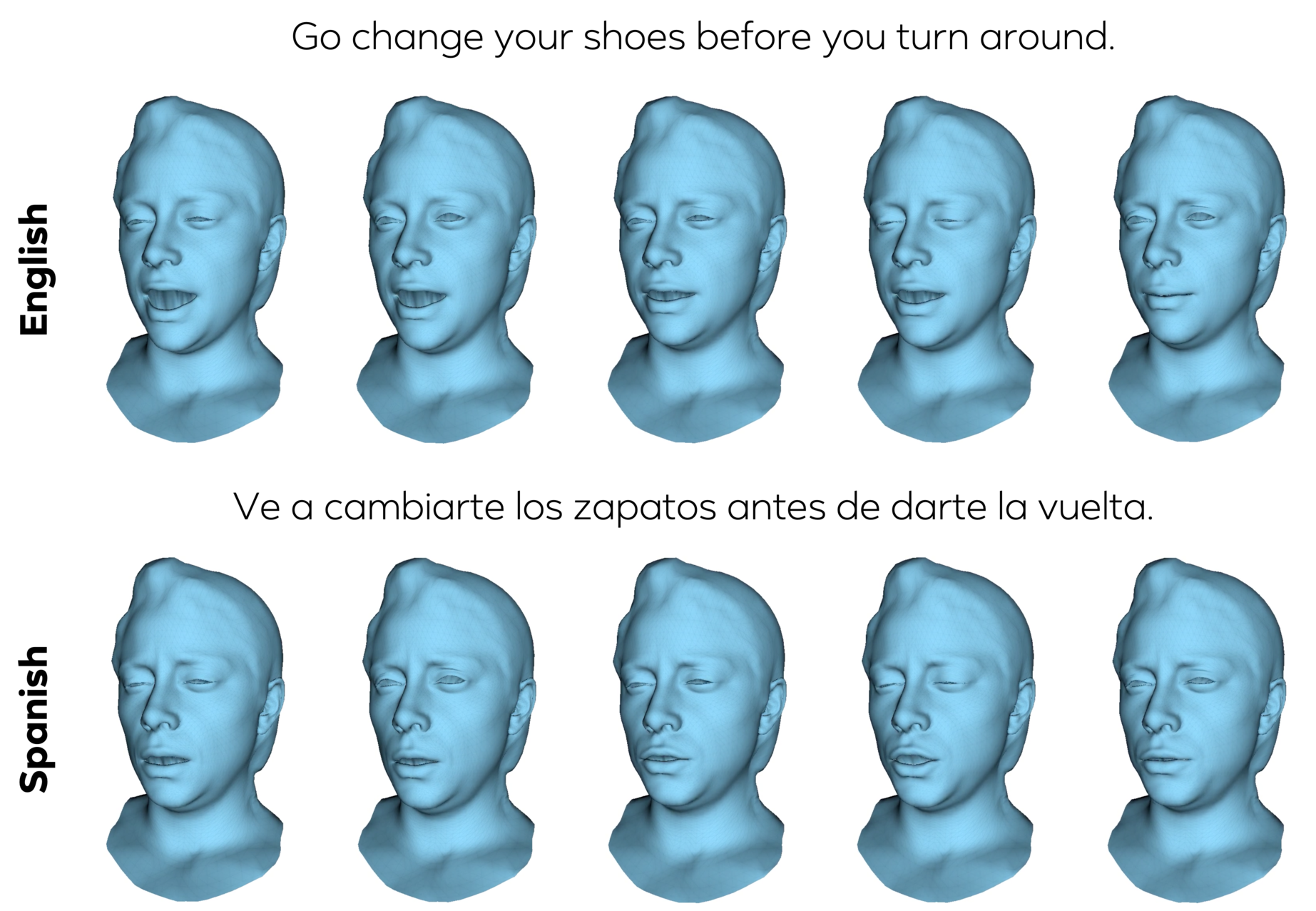}
    \caption{Dubbing. We re-synthesize an English sentence with a new Spanish audio snippet. Note how the lip shape is adjusted to the new audio but general upper face motion like eye closures are maintained.}
    \label{fig:dubbing}
\end{figure}
Usually, when dubbing videos, a speech translation that is fully consistent with the lip motion in the original language is not possible.
To overcome this, our system can be used to re-synthesize 3D face meshes with matching lip motion in another language while keeping upper face motion intact.
In this task, the original animated mesh and a new audio snippet are provided as input.
As before, we use the architecture in Figure~\ref{fig:overview} directly to re-synthesize lip motion in the new language.
Since the latent space is disentangled across modalities (see Figure~\ref{fig:latent_space} and Figure~\ref{fig:heatmesh}), lip motion will be adapted to the new audio snippet but the general upper face motion such as eye blinks are maintained from the original clip.
Figure~\ref{fig:dubbing} shows an example of this application.
See the supplemental video for further examples.
\section{Limitations}
While we have demonstrated state-of-the-art performance for audio driven facial animation both in terms of lip-sync as well as naturalness of the generated motion of the entire face, our approach is subject to a few limitations that can be addressed in follow-up work:
(1)
Our approach relies on audio inputs that extend $100$ms beyond the respective visual frame.
This leads to an inherent latency of $100$ms and prevents the use of our approach for online applications.
Please note, this `look ahead' is beneficial to achieve highest quality lip-sync, \eg, for sounds like `$/p/$' the lip closure can be modeled better.
%
%
(2)
Besides latency, our approach can not be run in real-time on low-cost commodity hardware such as a laptop CPU or virtual reality devices.
%
%
We believe that the computational cost can be drastically improved by further research.
(3)
If the face tracker fails to track certain parts of the face, \eg, if hair overlaps and occludes the eyebrows or eyes, we can not correctly learn the correlation of their motion to the audio signal.

\section{Conclusion}
We have presented a generic method for generating 3D facial animation from audio input alone.
A novel categorical latent space in combination with a cross-modality loss enables autoregressive generation of highly realistic animation.
Our approach has demonstrated highly accurate lip motion, while also synthesizing plausible motion of uncorrelated regions of the face.
It outperforms several baselines and obtains state-of-the-art quality.
%
%
We hope that our approach will be a stepping stone towards VR telepresence applications, as head mounted capture devices become smaller, thus complicating the regression of accurate lip and tongue motion from oblique camera viewpoints.

{\small
\bibliographystyle{ieee_fullname}
\bibliography{references}
}

\end{document}